\documentclass[letterpaper, 10pt, conference]{ieeeconf}  

\IEEEoverridecommandlockouts                              

\usepackage{cite}
\usepackage{bbm}
\usepackage{multirow}
\usepackage{graphicx} 
\usepackage{booktabs}
\usepackage[namelimits]{amsmath} 
\usepackage{amssymb}             
\usepackage{amsfonts}    
\usepackage{dsfont}
\usepackage{mathrsfs} 
\usepackage{lipsum}
\usepackage{url}
\usepackage{xcolor}
\usepackage{algorithm}
\usepackage{algpseudocode}
\usepackage{hyperref}
\usepackage[capitalize]{cleveref}
\hypersetup{
    colorlinks=true,
    linkcolor=blue,
    filecolor=magenta,      
    urlcolor=cyan
}

\title{\LARGE \bf
Gen-Drive: Enhancing Diffusion Generative Driving Policies with Reward Modeling and Reinforcement Learning Fine-tuning
}

\author{Zhiyu Huang$^{1}$, Xinshuo Weng$^{2}$, Maximilian Igl$^{2}$, Yuxiao Chen$^{2}$, \\ Yulong Cao$^{2}$, Boris Ivanovic$^{2}$, Marco Pavone$^{2,3}$, Chen Lv$^{1}$
\thanks{$^1$School of Mechanical and Aerospace Engineering, Nanyang Technological University, Singapore. E-mail: {\tt zhiyu001@e.ntu.edu.sg, lyuchen@ntu.edu.sg}}%
\thanks{$^{2}$NVIDIA Research, NVIDIA Corporation, Santa Clara, CA, USA. E-mail: {\tt \{xweng, migl, yuxiaoc, yulongc, bivanovic, mpavone\}@nvidia.com
}}%
\thanks{$^3$Department of Aeronautics and Astronautics, Stanford University, USA.}
}

\begin{document}

\maketitle
\thispagestyle{empty}
\pagestyle{empty}

\begin{abstract}
Autonomous driving necessitates the ability to reason about future interactions between traffic agents and to make informed evaluations for planning. This paper introduces the \textit{Gen-Drive} framework, which shifts from the traditional prediction and deterministic planning framework to a generation-then-evaluation planning paradigm. The framework employs a behavior diffusion model as a scene generator to produce diverse possible future scenarios, thereby enhancing the capability for joint interaction reasoning. To facilitate decision-making, we propose a scene evaluator (reward) model, trained with pairwise preference data collected through VLM assistance, thereby reducing human workload and enhancing scalability. Furthermore, we utilize an RL fine-tuning framework to improve the generation quality of the diffusion model, rendering it more effective for planning tasks. We conduct training and closed-loop planning tests on the nuPlan dataset, and the results demonstrate that employing such a generation-then-evaluation strategy outperforms other learning-based approaches. Additionally, the fine-tuned generative driving policy shows significant enhancements in planning performance. We further demonstrate that utilizing our learned reward model for evaluation or RL fine-tuning leads to better planning performance compared to relying on human-designed rewards. Project website: \url{https://mczhi.github.io/GenDrive}.
\end{abstract}

\section{Introduction}

Navigating complex environments requires autonomous driving agents to adeptly anticipate future scenarios (e.g., the behaviors of other agents) while making informed decisions \cite{huang2023gameformer, chen2023categorical}. Conventional predictive and deterministic planning approaches often separate the prediction and planning processes \cite{10563192, chen2023interactive}, which isolates the ego vehicle from the social context and often results in behaviors that do not comply with social driving norms. Although integrated prediction-planning frameworks \cite{huang2023differentiable, karkus2023diffstack, huang2023conditional, huang2024dtpp, ivanovic2020mats} have been proposed to address this issue, they still rely on deterministic planning, which poses challenges in addressing the uncertainties, multi-modality, and mutual interactive dynamics of agent behaviors. To overcome these challenges, we propose the adoption of generation-evaluation methods for the planning task. The key idea is that our approach integrates the ego agent into the social interaction context, generates a range of possible outcomes for all agents in the entire scene, and employs a learned scene evaluator to guide decision-making. While generative models, particularly diffusion models, have seen extensive use in simulation and prediction tasks in autonomous driving \cite{huang2024versatile, ctg, zhong2023language, jiang2023motiondiffuser, xu2023diffscene}, their application in decision-making tasks has been relatively limited. 

\begin{figure}[t]
    \centering
    \includegraphics[width=0.93\linewidth]{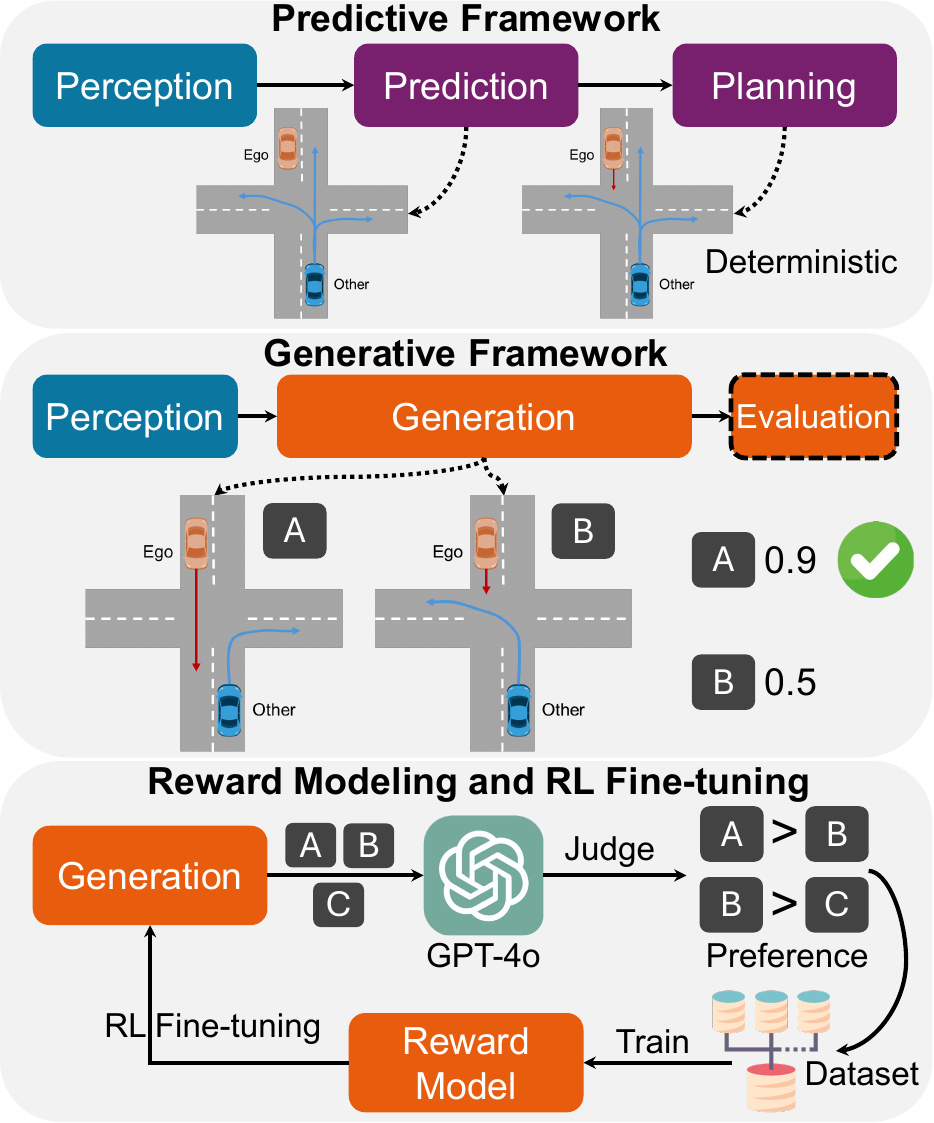}
    \caption{\textbf{\textit{Gen-Drive}} represents a paradigm shift from conventional prediction and deterministic planning approaches to a generation-then-evaluation framework. In this framework, different joint future scenes for both the ego agent and other agents are generated, followed by a selection process through a scene evaluation (reward) model. To train an effective reward model, we sample generation outcomes and employ a VLM-assisted pipeline to curate a pairwise preference dataset. The trained reward model can be utilized to make informed decisions and fine-tune the generation model via RL, further enhancing its planning performance.}
    \label{fig:1}
    \vspace{-0.6cm}
\end{figure}

Two primary limitations hinder the application of generative models in the planning task. First, it is complicated to evaluate generated scenarios and select the optimal one for decision-making that aligns with human expectations and values. To resolve this, we introduce a scene evaluation (reward) model trained on preference data derived from VLM feedback, enabling better decision-making. Second, unlike simulation or scenario generation tasks that benefit from a diversity of samples, planning with generative models requires producing more probable future scenarios with fewer samples to minimize computational overhead and runtime delays. We address this by introducing a reinforcement learning (RL) fine-tuning framework \cite{uehara2024understanding} that enhances the quality of diffusion generation based on the obtained reward model. The results demonstrate significant improvements in the model's planning performance after fine-tuning.

In this paper, we introduce \textbf{\textit{Gen-Drive}}, a diffusion generative driving policy, along with its training framework. The policy model comprises a query-centric scene context encoder in vector space \cite{zhou2023query}, a diffusion-based scene generator, and a scene evaluator to assess the quality of the generated scenes for planning. The training process involves three stages. Initially, we utilize a large volume of real-world driving data to train the base diffusion model, consisting of the scene encoder and generator. Subsequently, we curate a dataset of pairwise preference data on scenarios generated by the base diffusion model, employing a hybrid labeling pipeline assisted by vision language models (VLMs), and then train a scene evaluator based on the dataset. Finally, we employ reinforcement learning from AI feedback \cite{cao2023reinforcement} to fine-tune the diffusion generator, enhancing its efficacy in planning tasks. The comprehensive \textbf{\textit{Gen-Drive}} framework and its training process are illustrated in \cref{fig:1}. The contributions of this paper are summarized as follows: 
\begin{enumerate}
\item We develop a multi-agent trajectory diffusion model to reason about the interactions among all agents and generate diverse, scene-consistent future scenarios.
\item We train a reward model from a curated VLM-feedback preference dataset that is able to evaluate the goodness of the scenarios generated by the model.
\item We build an RL fine-tuning pipeline to enhance the performance of the diffusion driving policy based on the learned reward model.
\end{enumerate}
The base model is trained using the nuPlan dataset, and evaluated on the nuPlan closed-loop planning benchmark \cite{karnchanachari2024towards}. The results reveal that our diffusion driving policy achieves favorable performance, particularly after fine-tuning.

\section{Related Work}
\subsection{Prediction and Planning for Autonomous Driving}
Prediction-planning frameworks form the cornerstone of decision-making in autonomous driving, which predicts the behaviors of other agents and then plans for the ego agent. Recent advancements in deep learning have significantly enhanced the accuracy of prediction models \cite{huang2022multi, seff2023motionlm, shi2024mtr++}. Nonetheless, simply enhancing prediction accuracy may not necessarily translate to better planning performance \cite{huang2024dtpp, tran2023truly}. To address this, integrated prediction and planning (IPP) methods have been proposed \cite{hagedorn2023rethinking, karkus2023diffstack, huang2023differentiable, huang2024dtpp}, aiming to optimize overall planning performance. Although IPP approaches can directly enhance planning performance, they still isolate the ego vehicle from social interaction contexts and struggle to handle the inherent multi-modality of the prediction results. Moreover, the complexity of designing IPP methods and their complicated training requirements present challenges to their practical deployment. Consequently, we propose a shift from predictive to generative approaches, by developing a model capable of generating diverse future scenarios for all agents, including the ego vehicle. This is intended to ensure scene consistency, better capture interaction dynamics, and enhance final planning performance.

\subsection{Reward Modeling for Autonomous Driving}
Reward modeling or scene/planning evaluation is a crucial yet challenging task. Traditional evaluations primarily rely on metrics and functions crafted by humans, such as nuPlan score \cite{karnchanachari2024towards} and Predictive Driving Model (PDM) score  \cite{dauner2024navsim}. These metrics, while useful, may not reflect human values accurately across different scenarios and can result in planning models trained with these rewards diverging from human-like behaviors. A promising direction of reward modeling is to learn from human driving data using inverse reinforcement learning (IRL) \cite{huang2021driving, huang2024dtpp, huang2023conditional}. However, it requires assumptions about the structure of the reward function that may not reflect actual human preferences under varying conditions. Recently, leveraging pairwise human preference data to train reward models has gained popularity \cite{kaufmann2023survey}, which can further be applied to fine-tune generative models \cite{cao2023reinforcement}. This approach shows promise in delivering human-aligned evaluations and enhancing human-like driving capabilities.

\subsection{Generative Models for Autonomous Driving}
Generative models, such as diffusion and auto-regressive Transformer models, have been increasingly utilized in traffic simulation and trajectory prediction tasks \cite{zhou2024behaviorgpt, wu2024smart, hu2024solving, philion2023trajeglish, guo2023scenedm, huang2024versatile, seff2023motionlm, niedoba2024diffusion}. These models excel in capturing complex and multi-modal distributions of multi-agent joint behaviors, leading to exceptional performance in scenario generation performance \cite{zhou2024behaviorgpt, wu2024smart, huang2024versatile, niedoba2024diffusion}. Although auto-regressive generation models, such as BehaviorGPT \cite{zhou2024behaviorgpt} and SMART \cite{wu2024smart}, are adept at generating interactive behaviors, they fall short in diversity and controllability compared to diffusion models. Consequently, diffusion models represent a promising alternative. Notable examples in diffusion-based scenario generation include CTG++ \cite{zhong2023language} and VBD \cite{huang2024versatile}, which form the foundation of our framework. A particular application of diffusion models in planning is Diffusion-ES \cite{yang2024diffusion}, but it is mainly used as a trajectory optimization method for the ego agent instead of generating multi-agent future scenarios. In our proposed \textbf{\textit{Gen-Drive}} model, we aim to improve planning performance with proper reward modeling and to refine the diffusion model for planning. Training diffusion models with RL, previously successful in image generation \cite{zhang2024large, black2023training}, is employed to improve the model's generation quality by directing outcomes towards high-rewarding plans/scenarios and reducing computational cost.

\section{Method}

\begin{figure*}[htp]
    \centering
    \includegraphics[width=\linewidth]{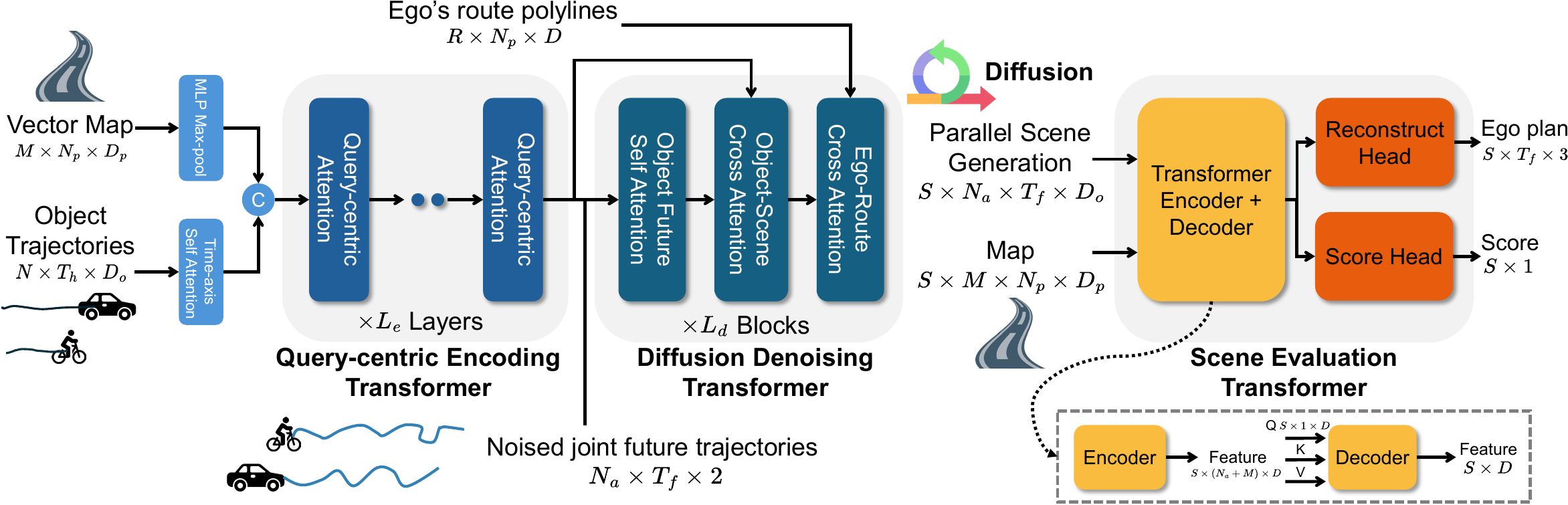}
    \caption{Neural network structure of the \textbf{\textit{Gen-Drive}} model. The query-centric encoding Transformer encodes all scene elements in local coordinates while preserving relative information in attention calculations. The diffusion denoising Transformer comprises multiple attention blocks that iteratively attend to noised object futures, future-scene, and ego-route interactions. During diffusion generation,  different scenes can be produced in parallel and then be fed into the scene evaluation Transformer. The evaluation model utilizes a Transformer encoder-decoder to fuse information from the future scene and map, and two MLP heads are used to reconstruct the ego plan and output a score for the scene/plan. }
    \label{fig:2}
    \vspace{-0.5cm}
\end{figure*}

We employ a generative (diffusion) model to replace the prediction-planning models in the conventional paradigm. The critical distinction is that the ego agent is not isolated from the scene; rather, it is considered an integral part, with all agents' behaviors being deeply interdependent. To leverage this generative model for planning, we design a scene evaluation (reward) model. This model is trained using a curated dataset of pairwise human preferences, enabling it to directly score the generated scenarios (plans) and facilitating the selection of optimal and contextually aligned decisions. Furthermore, we utilize the reward model to fine-tune the diffusion generation process, steering it towards generating high-rewarding plans. The RL fine-tuning step can enhance overall planning performance and reduce the need for extensive sampling. The neural network structure of the \textbf{\textit{Gen-Drive}} model is illustrated in \cref{fig:2}.

\subsection{Scene Generator}
For an initial driving scene at the current timestep, we consider $N$ objects (including the ego) and $M$ map elements, tracking the historical trajectories of these objects over $T_h$ timesteps. The current scene inputs to the encoder consist of object trajectories $\mathcal{O} \in \mathbb{R}^{N \times T_h \times D_o}$ and map polylines $\mathcal{M} \in \mathbb{R}^{M \times N_p \times D_p}$, where $N_p$ is the number of waypoints and $D_o$ and $D_p$ are the dimensional features of each point.  

\textbf{Encoder}.
The current scene inputs are initially encoded through a time-axis self-attention Transformer layer for the object trajectories, yielding $\mathcal{O}^e \in \mathbb{R}^{N \times D}$, and through an MLP with max-pooling for the map data, resulting in $\mathcal{M}^e \in \mathbb{R}^{M \times D}$. They are concatenated to form an initial encoding. We employ a query-centric Transformer encoder \cite{shi2024mtr++, zhou2023query, huang2024versatile} to fuse the features of the scene elements and produce a comprehensive scene condition encoding $\mathcal{C} \in \mathbb{R}^{(N+M) \times D}$.

\textbf{Denoiser}.
The diffusion process operates in the joint action space $\mathbf{a}$ of all objects of interest $N_a$, and action consists of acceleration and yaw rate. The noise is directly added to the action sequence. Given the noise input $\mathbf{a}_{k} \in \mathbb{R}^{N_a \times T_f \times 2}$, where $k$ is the noise level, $T_f$ is the future timesteps, as well as the scene condition $\mathcal{C}$, we employ a denoising Transformer with self-attention and cross-attention layers to predict the denoised action sequence $\mathbf{\hat a}_0$. For the ego agent, additional route information is provided and an additional cross-attention layer is employed to model ego-route relations. Further details on the diffusion process and model structure can be found in our previous work \cite{huang2024versatile}.

\textbf{Generation}.
Future scenarios (joint object actions) are generated starting from random Gaussian noise $\mathbf{a}_{K} \sim \mathcal{N}(0, \mathbf{I})$, where $K$ is the total number of diffusion steps. Subsequently, each diffusion step $k$ involves sampling from the transition dynamics specified below \cite{ho2020denoising, nichol2021improved, janner2022planning}:
\begin{equation}
\label{mean}
\mu_{k} :=  \frac{\sqrt{\bar \alpha_{k-1}} \beta_k}{1 - \bar \alpha_k} {\mathbf{\hat a}_0} + \frac{\sqrt{\alpha_k} (1 - \bar \alpha_{k-1})}{1 - \bar \alpha_k} \mathbf{a}_k,    
\end{equation}
\begin{equation}
\label{reverse}
p(\mathbf{a}_{k-1} | \mathbf{a}_{k}) = \mathcal{N} \left(\mathbf{a}_{k-1}; \mu_{k}, \frac{1 - \bar \alpha_{k-1}}{1 - \bar \alpha_k} \beta_k \mathbf{I} \right),    
\end{equation}
where $\alpha_k$, $\bar \alpha_k$, and $\beta_k$ are derived from a predetermined noise schedule. By iteratively reversing the diffusion step, we obtain the final denoised joint action output $\mathbf{a}_0$. Subsequently, states are derived using a dynamics model $f$ to translate object actions into states $\mathbf{x}_0 = f(\mathbf{a}_0)$. The state encompasses $x/y$ coordinates, heading, and velocity of the object.

\subsection{Scene Evaluator} 
The scene evaluator takes as input $S$ future scenarios produced by the diffusion generator, which can be generated in parallel by initiating from a batch of Gaussian noise. These generated scenarios are structured as $\mathcal{S} \in \mathbb{R}^{S \times N_a \times T_f \times D_o}$, and another input to the evaluator is the vector map $\mathcal{M}$. These future scenes are encoded using a query-centric Transformer encoder, similar to the encoding of historical scenes, resulting in a scene feature representation $\mathcal{S}^e \in \mathbb{R}^{S \times (N_a + M) \times D}$. Subsequently, we employ the ego agent's future or planning encoding $\mathcal{A}^e \in \mathbb{R}^{S \times 1 \times D}$ extracted from the scene encoding $\mathcal{S}^e$ as query, and the scene encoding as key and value in a Transformer decoder, to derive the planning-centric feature of the future scenes $\mathcal{A} \in \mathbb{R}^{S \times D}$. Note that the Transformer decoder attends to the $(N_a+M)$ elements within each scene individually. Two MLP heads are appended to this feature tensor to reconstruct the planning trajectories of the ego agent and output scores for different generated scenes (planning trajectories), respectively. The ego planning reconstruction head is added as an auxiliary task to enhance stability and effectiveness.

\subsection{Training Base Diffusion Model}
The base diffusion model is trained to recover clean trajectories from noised joint trajectory inputs under various noise levels and scene conditions. At each training step, a noise level $k$ and Gaussian noise are sampled to perturb the original action trajectories. Since the model predicts scene-level joint trajectories, all object trajectories are affected by the same noise level. The training loss function for the base diffusion model can be formulated as: 
\begin{equation}
\mathcal{L}_{\mathcal{G}} = \mathbb{E}_{(\mathcal{O}, \mathcal{M}) \sim D, k \sim \mathcal{U}(0, K)}
    \left[ \mathcal{SL}_1 \left( f\left( \mathcal{G} (\mathcal{O}, \mathcal{M}, \mathbf{a}_k, k) \right)  - \mathbf{x} \right) \right],
\end{equation}
where $D$ is the dataset, $\mathcal{SL}_1$ is the Smooth L1 loss, $f$ is the dynamics model, $\mathbf{x}$ is the ground-truth future states of the objects. $\mathcal{G}$ denotes the diffusion model that predicts clean trajectories given noised trajectories $\mathbf{a}_k$ and scene conditions.

\subsection{Training Reward Model}
\textbf{Pairwise preference data collection}.
To build an effective reward model, it is essential to curate a comprehensive dataset. One approach involves utilizing human-designed metrics such as the PDM score \cite{dauner2024navsim}. However, relying on such metrics presents significant limitations, as they may not accurately reflect actual human values across diverse scenarios. Additionally, accurately labeling scenes with reward values is challenging even for human evaluators. Alternatively, we can engage human annotators to perform pairwise comparisons, determining which scenarios more align with human preferences. Nevertheless, curating a large-scale reward dataset imposes a substantial workload on human annotators. To address this, we employ VLMs to enhance the efficiency and scalability of the process.

\begin{figure}[htp]
    \centering
    \includegraphics[width=0.98\linewidth]{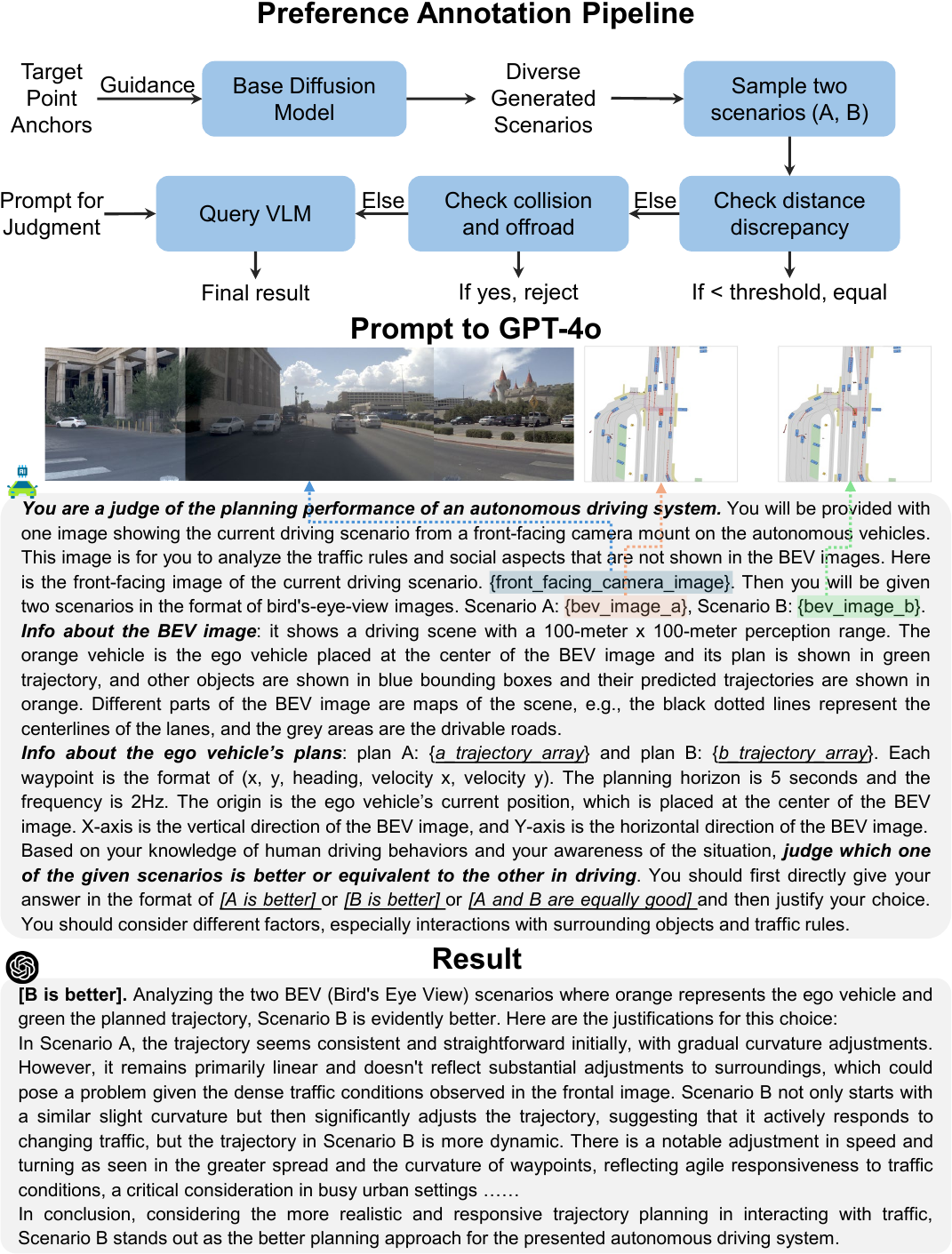}
    \caption{Pipeline for collecting planning preference data using VLM.}
    \label{fig:3a}
     \vspace{-0.4cm}
\end{figure}

The VLM-assisted reward labeling pipeline is illustrated in \cref{fig:3a}. To increase the diversity of planning trajectories, we first extract 32 5-second anchor goals from data utilizing the K-means algorithm and employ guided diffusion \cite{huang2024versatile, zhong2023language} to generate 32 diverse planning trajectories for the ego agent along with reactive behaviors of other objects in the scene by the model. Subsequently, we conduct pairwise sampling of these scenarios. We first compute discrepancies between the planned trajectories, and then check collisions and off-road to filter out obvious failure cases. If these measures are insufficient for distinction, we utilize GPT-4o to provide a conclusive evaluation. As illustrated in \cref{fig:3a}, GPT-4o provides reasonable evaluations of the two generated scenarios based on the current scene context.

\begin{figure}[htp]
    \centering
    \includegraphics[width=0.825\linewidth]{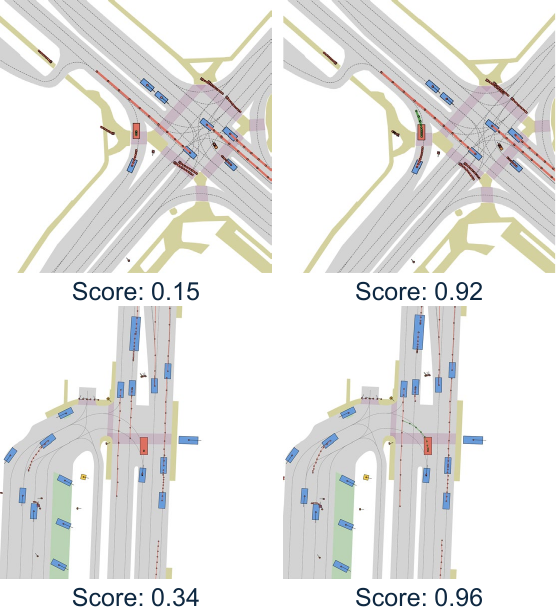}
    \caption{Examples of scene evaluation (reward model) outputs; plans that ensure safety, interactivity, and making progress receive higher scores.}
    \label{fig:3b}
     \vspace{-0.5cm}
\end{figure}

\textbf{Training process}.
At each training step, we sample a batch of pairwise comparison results, i.e., accepted (positive) scenes $\mathcal{S}_a$ and rejected (negative) scenes $\mathcal{S}_r$ from the same initial conditions. The loss function for training the scene evaluation model is formulated as follows \cite{cao2023reinforcement, abramson2022improving}:
\begin{equation}
\mathcal{L}_{\mathcal{R}} = \mathbb{E}_{(\mathcal{S}_a, \mathcal{S}_r) \sim D_r}
    \left[ -\log \sigma \left( \mathcal{R}(\mathcal{S}_a) - \mathcal{R}(\mathcal{S}_r) \right) \right],
\end{equation}
where $D_r$ is the pairwise preference reward dataset, and $\mathcal{R}$ denotes the reward model that predicts the score of the generated scene. Note that the reconstruction loss is omitted for brevity. Some examples of the reward model outputs are illustrated in \cref{fig:3b}, and the results demonstrate that our trained reward model produces reasonable scores for the generated plans and scenes. 

\subsection{Fine-tuning Generation Model}
To enhance the efficacy of diffusion generation for planning tasks, we propose fine-tuning the diffusion model using the trained reward model and RL. We can formulate the diffusion denoising process as a multi-step MDP \cite{black2023training, uehara2024understanding}, where the denoiser functions as a policy conditioned on the noise input at each step. The trajectory comprises $K$ timesteps, with a reward signal emitted at the end of the diffusion process. The RL objective is to maximize the cumulative reward along the trajectory, and we can utilize denoising diffusion policy optimization (DDPO) \cite{black2023training} to refine the generative policy. The fine-tuning loss is formulated as: 
\begin{equation}
\label{rlft}
\mathcal{L}_F = \sum_{k=K}^{0} \Big[ \mathcal{L}_{policy} + \alpha \mathcal{SL}_1 (\mathbf{\hat x}_0 - \mathbf{x}) \Big],
\end{equation}
\begin{equation}
\begin{split}
\mathcal{L}_{policy} = - \min \Big\{ r(\mathbf{x}_0) \frac{p(\mathbf{x}^e_{k-1} |\mathbf{x}^e_{k} ; \theta)}{p(\mathbf{x}^e_{k-1} | \mathbf{x}^e_{k} ; \theta_{old})}, \\
 r(\mathbf{x}_0) \ \text{Clip} \Big( \frac{p(\mathbf{x}^e_{k-1} |\mathbf{x}^e_{k} ; \theta)}{p(\mathbf{x}^e_{k-1} | \mathbf{x}^e_{k} ; \theta_{old})}, 1 - \epsilon, 1 + \epsilon \Big)  \Big\},    
\end{split}
\end{equation}
where $\mathbf{\hat x}_0$ is the denoised state trajectories for all objects, $\mathbf{x}$ is the ground-truth trajectories, $\mathbf{x}^e_k$ is the state trajectory for the ego agent; $r(\mathbf{x}_0)$ is the reward of the generated scene; $\alpha$ is a regularization parameter, and $\epsilon$ is the clipping parameter. A regression term is added to ensure the denoised trajectory does not deviate from the original one.

Note that the fine-tuning loss is accumulated over the entire diffusion trajectory, and only the denoiser is learnable while the encoder is fixed during fine-tuning. The RL fine-tuning algorithm with DDPO is illustrated in \cref{algo1}.
 
\begin{algorithm}
    \caption{Fine-tuning of generation with DDPO}
    \begin{algorithmic}[1]
    \Require Pre-trained generation model $\mathcal{G}_{\theta_{pre}}$, trained reward model $\mathcal{R}$, sample size $m$, learning rate $\gamma$, fine-tuning steps $T$, update iterations $I$
    \State \textbf{Initialize:} $\mathcal{G}_{\theta_1} \gets \mathcal{G}_{\theta_{pre}}$
    \For{$t \gets 1 \text{ to } T$} 
        \State Generate $m$ samples of generation sequence $\{\mathbf{a}^{(i)}_k \}_{k=0}^{K}$ from the current diffusion model $\mathcal{G}_{\theta}$
        \State Collect $m$ samples of scene reward $\{ r^{(i)} \}$ using $\mathcal{R}$ 
        \State Normalize rewards $\{ r^{(i)} \} = \frac{r^{(i)} - mean(\{ r^{(i)} \})}{std(\{ r^{(i)} \})}$
        \For{$i \gets 1 \text{ to } I$}
            \State Compute loss function $\mathcal{J}(\theta) = \frac{1}{m} \sum_{M} \mathcal{L}_F$ \Comment{Refer to \cref{rlft}}
            \State Update $\theta_{t+1} \gets \theta_t - \gamma \nabla_{\theta} \mathcal{J}(\theta)$ 
        \EndFor
    \EndFor
    \State \textbf{Output:} Fine-tuned generation model $\mathcal{G}_{\theta_{T}}$
    \end{algorithmic}
    \label{algo1}
\end{algorithm}
\vspace{-0.3cm}

\subsection{Implementation Details}
\textbf{Model Parameters}. 
The scene context consists of $N=100$ objects, each with a 2-second historical trajectory, sampled at 0.5-second intervals ($T_h=4$). The vector map contains $M=350$ elements, with each comprising $N_p=20$ waypoints, and additional $R=30$ route polylines for the ego vehicle. The base diffusion model consists of 6 encoding layers with query-centric attention and 6 Transformer decoding layers, with a hidden dimension of $D=256$. The model generates future scenes for $N_a=50$ objects closest to the ego agent over a 5-second horizon, with 0.5-second intervals ($T_f=10$). We employ $K=10$ diffusion steps and a cosine noise schedule. The scene evaluator comprises 3 query-centric attention layers in the Transformer encoder and 3 cross-attention layers in the Transformer decoder. 

\textbf{Training Pipeline}.
For training the base diffusion model, we employ a batch size of 16 per GPU and 20 training epochs. AdamW optimizer is used, starting with a learning rate of 2e-4, which decays 0.95 every 1k steps. For the reward model, we collect 5k scenarios, each with 50 pairwise comparisons, and train it using a batch size of 32 per GPU across 50 epochs. In the fine-tuning phase, the parameters are set: regularization $\alpha=10$, clip parameter $\epsilon=0.01$, sample size $m=32$ per GPU, learning rate $\gamma=1e-5$, fine-tuning steps $T=1000$, and update iterations per step $I=5$. All training jobs are conducted on 8 NVIDIA A100 GPUs.

\section{Experiments}
\subsection{Experimental Setup}
\textbf{Dataset}.
We employ the OpenScene dataset \cite{dauner2024navsim}, which is a compact subset of the nuPlan dataset sampled at 2Hz. Only the vectorized data, including trajectories and maps, were employed to train the base behavior diffusion model. The training set comprises 450k segments, each comprising a 2-second history and a 5-second future. No data augmentation methods are used. For reward model training and generation fine-tuning, we select 5k scenarios from the test split.

\textbf{Closed-loop Planning Test}.
We test the model's performance on the nuPlan closed-loop planning task with non-reactive agents \cite{karnchanachari2024towards}. The planning frequency is set at 1 Hz, and the trajectory controller is set to iLQR. Evaluation metrics include the overall planning score, collision score, and progress score (higher scores indicate better performance). We adopt the reduced Val14 set as the benchmark \cite{dauner2023parting}.

\subsection{Main Results}
\begin{figure*}[htp]
    \centering
    \includegraphics[width=0.914\linewidth]{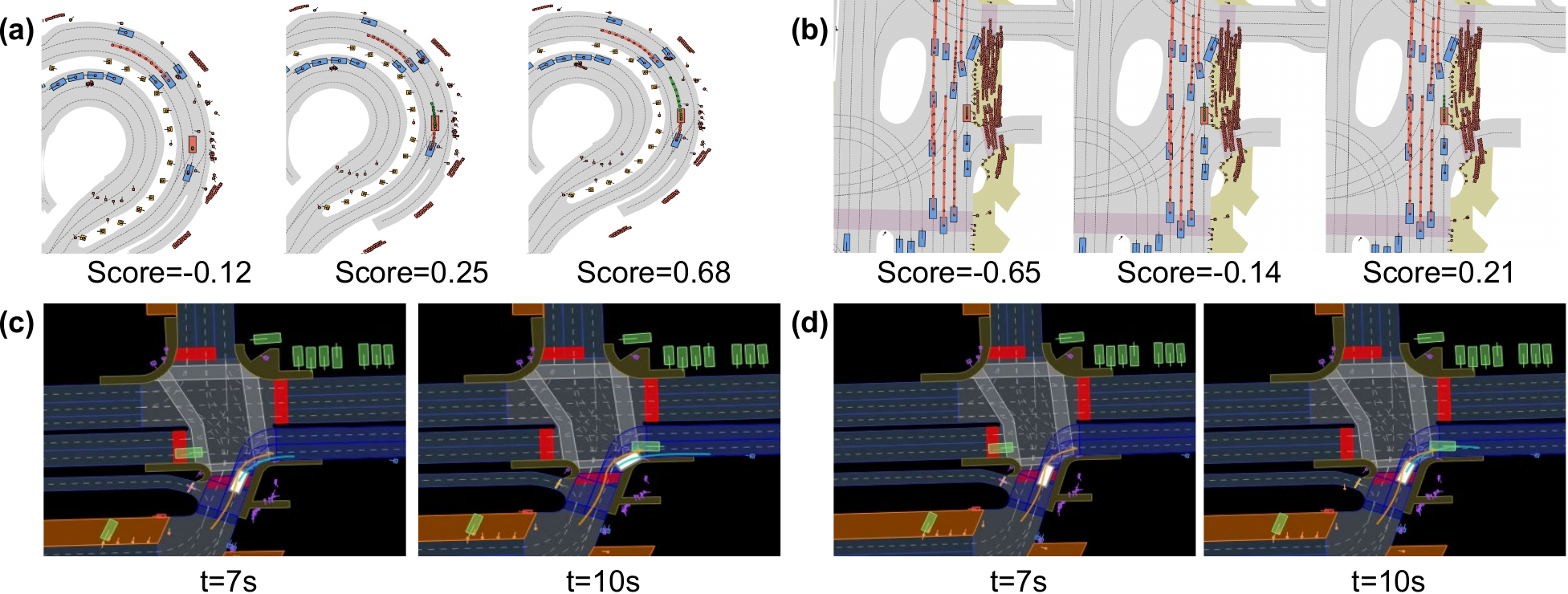}
    \caption{Illustration of the planning process. (a) and (b) show two scenarios where our generative planner produces diverse plans and interaction outcomes. (c) and (d) compare the performance of the base policy with the fine-tuned policy, showing that the fine-tuned one learns to yield and avoid off-road plans. }
    \label{fig:4}
    \vspace{-0.5cm}
\end{figure*}

The closed-loop planning results for different models are presented in \cref{table_1}. Furthermore, \cref{fig:4} illustrates some typical scenarios from the generation process, and the fine-tuned policy shows better planning performance. Closed-loop planning results can be found on the \href{https://mczhi.github.io/GenDrive/}{project website}. 

\begin{table}[htp]
\caption{Closed-loop Planning Results on nuPlan reduced Val14 Benchmark}
\label{table_1}
\begin{center}
\begin{tabular}{@{}l|ccc@{}}
\toprule
Method                              & Score             & Collision             & Progress \\ \midrule
PDM-Closed \cite{dauner2023parting} & \textbf{0.9121}   & \textbf{0.9701}       & \textbf{0.9268}         \\
GameFormer \cite{huang2023gameformer}& 0.8376           & 0.9473                & 0.8812         \\
DTPP \cite{huang2024dtpp}           & 0.8243            & 0.9292                & 0.8323          \\
PlanTF \cite{cheng2024rethinking}   & 0.8366            & 0.9402                & 0.9267   \\
\midrule
Gen-Drive (Single)                  & 0.8260            & 0.9160                & 0.8435     \\ 
Gen-Drive (Multiple + PDM)          & 0.8087            & 0.9253                & 0.8587         \\ 
Gen-Drive (Multiple + Learned)      & 0.8512            & 0.9365                & 0.8664         \\ 
Gen-Drive (Fine-tuned, Single)      & 0.8633            & 0.9414                & 0.9012         \\ 
Gen-Drive (Fine-tuned, Multiple)    & \textbf{0.8753}   & \textbf{0.9572}       & \textbf{0.8994} \\ 
Gen-Drive (PDM fine-tuned, Single)  & 0.8550             & 0.9455              & 0.8655       \\   \bottomrule
\end{tabular}
\end{center}
\vspace{-0.3cm}
\end{table}

\textbf{Generation and evaluation outperforms single-sample inference}. 
The multiple-sample planning method generates 16 scenes in parallel through batch processing and selects the optimal scenario using a learned reward model. This method enhances the diversity of generated plans, thereby improving the overall planning score. Additionally, the generative planner with our learned reward model outperforms the PDM score-based evaluator in planning. The average inference time for single-sample planning is 282.5 ms and 483.6 ms for multiple-sample planning on an RTX 4090 GPU, maintaining real-time performance requirements.

\textbf{Fine-tuning enhances performance}.
Planning efficacy still largely depends on generation quality, and we demonstrate that RL fine-tuning can substantially enhance quality and performance. Notably, even with the single-sample method, the overall planning score of the fine-tuned policy outperforms the multiple-sample approach without fine-tuning. Moreover, using our learned reward model for fine-tuning performs better than using the PDM-based scorer.

\textbf{Comparison with predictive methods}. 
Compared to learning-based predictive planners (PlanTF, GameFormer, and DTPP), our model exhibits superior performance by using a generation-then-evaluation method. However, the PDM-Closed planner, which utilizes a rule-based trajectory generator and scorer, achieves the highest scores. It is important to note that it is optimized for nuPlan metrics, which may lack human likeness and adaptability to real-world scenarios. 

\subsection{Ablation Studies}
\textbf{Influence of the number of modeling objects}.
We investigate the effect of the number of objects in the scene generator, ranging from 1 (where only the ego vehicle's plan is generated) to 100, and present the outcomes in \cref{table_2}. We adjust the number of modeling objects in model training and employ a single-sample generation in testing. The results indicate that generating only the ego vehicle's plan results in inferior performance, primarily due to lack of movement in some cases. Conversely, an excessive number of modeling objects (e.g., 100) also results in diminished performance and runtime efficiency. Therefore, modeling 50 ego and surrounding objects performs best while maintaining runtime efficiency.

\begin{table}[htp]
\caption{Influence of the number of modeling objects}
\label{table_2}
\begin{center}
\begin{tabular}{@{}l|ccc@{}}
\toprule
Number of objects                   & Score             & Collision         & Progress \\ \midrule
1 (ego only)                        & 0.8118            & \textbf{0.9384}   & 0.7990        \\
20                                  & 0.8188            & 0.9160            & 0.8383         \\
50                                  & \textbf{0.8260}   & 0.9160            & \textbf{0.8435} \\
100                                 & 0.8079            & 0.9253            & 0.8166         \\   \bottomrule
\end{tabular}
\end{center}
\vspace{-0.6cm}
\end{table}

\textbf{Influence of number of fine-tuning steps}.
We examine the effect of training steps in the RL fine-tuning phase, employing a multiple-sample generation and scoring method in testing. The results shown in \cref{table_3} reveal that 1000 fine-tuning steps achieve the best planning metrics, beyond which the performance of the fine-tuned policy tends to decline. This is a common issue in RLHF frameworks, as the policy might exploit the reward function and generate unreasonable behaviors. Therefore, we confine the RL fine-tuning phase to 1000 steps to prevent performance degradation.

\begin{table}[h]
\caption{Influence of the number of RL fine-tuning steps}
\label{table_3}
\begin{center}
\begin{tabular}{@{}l|ccc@{}}
\toprule
Fine-tuning steps          & Score             & Collision         & Progress \\ \midrule
500                        & 0.8441            &  0.9354           & 0.8459         \\
1000                       &\textbf{0.8753}    & \textbf{0.9572}   & \textbf{0.8994}       \\
1500                       & 0.8457            & 0.9324            &  0.8431        \\
2000                       & 0.8123            & 0.9211            & 0.8345 \\ \bottomrule
\end{tabular}
\end{center}
\vspace{-0.6cm}
\end{table}

\section{Conclusions}
We propose the \textbf{\textit{Gen-Drive}} framework, marking a paradigm shift to generation-evaluation in decision-making for autonomous driving. The framework integrates a behavior diffusion model as the core scene generator to model multi-agent joint interactions, coupled with a scene evaluator learned using VLM-assisted preference data. Moreover, we apply RL fine-tuning to diffusion generation to further enhance its efficacy in planning tasks. Experimental results demonstrate the superior performance of our model with proper reward modeling compared to other learning-based planning methods and further enhancements through RL fine-tuning. Future research will aim to incorporate a perception module from sensor inputs within this generative framework to establish a fully end-to-end learnable driving system.


\bibliographystyle{IEEEtran}
\bibliography{IEEEexample}

\end{document}